\documentclass[sigconf,natbib=true,anonymous=false]{acmart}
\usepackage{graphicx}
\usepackage{csvsimple}
\usepackage{siunitx}

\AtBeginDocument{%
  \providecommand\BibTeX{{%
    \normalfont B\kern-0.5em{\scshape i\kern-0.25em b}\kern-0.8em\TeX}}}

\setcopyright{acmlicensed}
\copyrightyear{2024}
\acmYear{2024}
\acmDOI{XXXXXXX.XXXXXXX}

\acmConference[SIGIR '24]{In Proceedings of
the 47th International ACM SIGIR Conference on Research and Development
in Information Retrieval}{July 14--18,
  2024}{Washington, DC}
\acmISBN{978-1-4503-XXXX-X/18/06}

\begin{document}

%%
%% The "title" command has an optional parameter,
%% allowing the author to define a "short title" to be used in page headers.
\title{Investigating the Synergistic Effects of Dropout and Residual Connections on Language Model Training}

%%
%% The "author" command and its associated commands are used to define
%% the authors and their affiliations.
%% Of note is the shared affiliation of the first two authors, and the
%% "authornote" and "authornotemark" commands
%% used to denote shared contribution to the research.

\author{Qingyang Li}
\affiliation{%
  \institution{Drexel University}
  \streetaddress{3141 Chestnut St}
  \city{Philadelphia}
  \state{PA}
  \postcode{19104}
  \country{USA}}
\email{jl4533@drexel.edu}

\author{Weimao Ke}
\affiliation{%
  \institution{Drexel University}
  \streetaddress{3141 Chestnut St}
  \city{Philadelphia}
  \state{PA}
  \postcode{19104}
  \country{USA}}
\email{wk@drexel.edu}

%%
%% By default, the full list of authors will be used in the page
%% headers. Often, this list is too long, and will overlap
%% other information printed in the page headers. This command allows
%% the author to define a more concise list
%% of authors' names for this purpose.
\renewcommand{\shortauthors}{Li and Ke}

%%
%% The abstract is a short summary of the work to be presented in the
%% article.
\begin{abstract}
% TODO: Summarize the main goal, method, and findings of the paper (100-150 words).
This paper examines the pivotal role of dropout techniques in mitigating overfitting in language model training. It conducts a comprehensive investigation into the influence of variable dropout rates on both individual layers and residual connections within the context of language modeling. Our study conducts training of a decoder implementation on the classic Tiny Shakespeare data to examine the effects of the adjustments on training efficiency and validation error. Results not only confirm the benefits of dropout for regularization and residuals for convergence, but also reveal their interesting interactions. There exists an important trade-off between the depth of residual connections and the dropout on these connections for optimal deep neural network convergence and generalization.
\end{abstract}

%%
%% The code below is generated by the tool at http://dl.acm.org/ccs.cfm.
%% Please copy and paste the code instead of the example below.
%%
\begin{CCSXML}
<ccs2012>
   <concept>
       <concept_id>10010147.10010341.10010342.10010344</concept_id>
       <concept_desc>Computing methodologies~Model verification and validation</concept_desc>
       <concept_significance>500</concept_significance>
       </concept>
 </ccs2012>
\end{CCSXML}

\ccsdesc[500]{Computing methodologies~Model verification and validation}

% \ccsdesc[500]{Do Not Use This Code~Generate the Correct Terms for Your Paper}
% \ccsdesc[300]{Do Not Use This Code~Generate the Correct Terms for Your Paper}
% \ccsdesc{Do Not Use This Code~Generate the Correct Terms for Your Paper}
% \ccsdesc[100]{Do Not Use This Code~Generate the Correct Terms for Your Paper}

%%
%% Keywords. The author(s) should pick words that accurately describe
%% the work being presented. Separate the keywords with commas.
\keywords{Large Language Model, Generative Pretrained Transformer (GPT), Transformer Networks, Deep Learning}

% \received{1 February 2024}
% \received[revised]{12 March 2024}
% \received[accepted]{5 June 2024}

%%
%% This command processes the author and affiliation and title
%% information and builds the first part of the formatted document.
\maketitle

\section{Introduction}

% TODO: Introduce the problem area, significance of the research, and a brief overview of GPT+IR.
% TODO: State the paper's main contribution.
Research in Natural Language Processing (NLP) has evolved rather rapidly, where deep learning \cite{lecun2015deep} coupled with language modeling (LM) has become a cornerstone. Key challenges in LLMs using deep neural networks include overfitting and vanishing gradient descent, particularly as the network depth increases. Extensive research has been conducted on two major techniques to address these issues, namely connection dropouts and residual connections.

Dropout minimizes overfitting by randomly disabling neurons during training, alleviates over-reliance on specific features in the learning process and promotes more robust feature learning.\cite{srivastava2014dropout} %This technique effectively simulates training multiple networks, enhancing generalization. 
Residual connections support smoother model training of deeper networks by adding a layer’s output to that of subsequent layers.\cite{he2016deep} %This approach mitigates the vanishing gradient problem, maintaining better information flow throughout the network. 

This study explores the impact of varying dropout rates and residual connections in the Transformer architecture for language modeling. Using a decoder trained on the classic literature, we analyze how these architectural adjustments affect training convergence, validation errors, and generalizability. % The goal is to give insight into the interaction between the two mechanisms, determine optimal dropouts on layer connections as well as residual connections, and ultimately shed light on the balanced network topology enabling efficient convergence and generalizability.

\section{Background}

% TODO: Discuss existing work related to GPT, information retrieval, and their combination.
In the following sections, we discuss research in language modeling and deep learning, and delve into associated methodologies that we focus on in this study. 

% TODO: Introduce the concepts of GPT, IR, and why their integration is beneficial.
%  Background: 1 paragraph language models, 1 paragraph on transformer for LMs, 1 paragraph on dropout for regularization, and 1 paragraph on residual and its impact. Add related references in .bib and cite them.

\subsection{Language Models}
% TODO: 
% Discuss traditional information retrieval and retrieval-based approaches to question answering tasks
A language model is essentially a probability distribution model for natural language\cite{rosenfeld2000two}. It operates by calculating the likelihood of a sequence of words.% $w_1, w_2, \ldots, w_m$ where $m$ represents the number of words in the sequence. 
The model's function is to determine the probability of these words appearing in a specific sequence.%, denoted as $P(w_1, w_2, \ldots, w_m)$. 
This capability enables the model to predict the most probable word to occur next in a given context or to identify which word is more likely to appear based on the surrounding words.

%A Large Language Model (LLM) is a deep learning model, pre-trained on extensive datasets. Based on the latest Transformer architecture, it comprises an intricate architecture that includes both encoder and decoder components, each embedded with a self-attention mechanism. These components are adept at extracting semantic content from text and discerning intricate relationships among words (the meaning) and sequences (the context). This advanced structure enables an LLM to process and follow natural languages with remarkable efficiency and accuracy.

\subsection{Transformer For LMs}
% TODO: 
% Discuss traditional NLP approaches and language models
% Applications for information retrieval, summarization, and question answering

Transformers is a deep learning architecture and the backbone of the latest iteration of Large Language Modeling (LLMs). Introduced in the seminal paper "Attention Is All You Need" by Vaswani et al.\cite{vaswani2017attention}, transformers utilize the self-attention mechanism, which allows the model to weigh the significance of different parts of the input data independently. This feature enables the transformer to capture complex dependencies and contextual relationships in text, regardless of their distance in the sequence. Transformers consist of an encoder and a decoder, each containing multiple layers of self-attention and feed-forward neural networks. In the context of LLMs, transformers are trained on vast amounts of text data, learning to generate, interpret, and predict language tokens. 

\subsection{Dropout For Regularization}

Deep learning leverages neural networks to discern complex patterns within data. For enhanced learning of diverse patterns, deeper neural networks are often employed, thereby broadening the model's generalizability. However, in practical applications, excessively deep networks can lead to deteriorating performance, a phenomenon known as overfitting. This occurs when the network excessively focuses on the training set. To address overfitting, a mechanism called dropout, introduced by Srivastava et al. \cite{srivastava2014dropout}, has been implemented. Dropout, a regularization technique, functions by randomly deactivating (setting to zero) certain connections within the network during the training phase. This approach compels the network to adapt to functioning with a reduced number of units, thereby enhancing its robustness against overfitting. Dropout has proven effective in boosting the generalizability of neural networks and is now a common practice in the field.

\subsection{Residual Connection}

When training neural networks, the deeper the network, the more challenging it becomes for learning to converge. He et al. addressed this issue by introducing the concept of residual connections.\cite{he2016deep} These connections provide an alternative route for information to travel deeper into the network, effectively bypassing some layers.

Imagine a series of layers from layer $i$ to layer $i + n$, with $F$ representing the function of these layers. Let the input to layer $i$ be $x_i$. In a standard feed-forward approach, $x_i$ would sequentially pass through each of these layers, resulting in the output at layer $i + n$ being $y = F(x_i)$. Bypassing these layers of connections, the residual mechanism adds the input directly to the output, essentially making $y = x_i + F(x_i)$ (with linear layer normalizations). 

% In conventional feedforward neural networks, training deep architectures is often hindered by issues like exploding or vanishing gradients. However, the incorporation of residual connections has demonstrated a significant improvement on training these networks. 
By allowing information to bypass certain layers, these connections facilitate smoother information flow and gradient propagation, thus aiding in the convergence of deeper networks. 
% TODO
% Discuss potential gaps and what to be done/improved in future research. 
Extensive research has been done on mechanisms related to dropouts and residuals, we have yet to understand the interaction of the two and how deep neural network for language models can be further optimized with the interplay of these factors. 

\section{Methodology}
% TODO: Describe the GPT+IR system architecture.

Our approach involves systematically adjusting dropout rates for various combinations of attention (and MLP) and residual connections. The range of dropout rates combined with parameters on the residual connections allows us to explore the potential synergistic effects of these parameters on model training and performance. We hypothesize that specific combinations of dropout rates in direct and residual pathways can lead to accelerated convergence and lower validation errors. 

\subsection{Transformer Architecture}

We use a decoder of the transformer architecture to build a language model for next token prediction and sentence completion tasks. The process includes tokenization, embedding, positional embedding, and layer normalization, which we briefly discuss here: 

\begin{enumerate}
    \item Tokenization: Based on a character uni-gram model, the text is splitted into a sequence of individual characters. This is a rudimentary technique for language modeling, simple to study yet challenging to train given the amount of embiguity in each character. 
    \item Word Embedding: Once the text is tokenized, each token is transformed into a high-dimensional vector using an ebmedding layer. This embedding layer is typically learned during the training process. It maps each token to a dense vector that captures potential semantic and syntactic information. 
    \item Attention Mechanism: The attention--or self-attention--mechanism enables the model to factor in the importance of each token within the input sequence depending on their relative positions. In the decoder architecture, attention is focused on positions leading to the current token (to be predicted) during training, enabling the model to consider the preceding context of the sequence when making predictions. 
    \item Layer Normalization: This steps normalize the output of each layer to ensure consistent mean and variance across layers, leading to more stabilized learning and faster convergence. Likewise, batch normalization is also performed \cite{ba2016layer}. 
\end{enumerate}

\subsection{Dropout and Residual Connections}

With the above processes in place, we focus in this study two particular mechanisms shown to be crucial for the learning (back-propagation) of deep neural networks and regularization of the trained model: 

\begin{itemize}
    \item Dropout: Dropout is a regularization technique to prevent overfitting and improve generalizablity of the trained model. By randomly selecting a fraction of input and disregard their contribution, dropout helps improve model robustness and mitigates the problem of over-dependence on any single node. In the transformer architecture, dropout can be applied to various components including embeddings, attention layers, and other layers of the feed-forward network. 
    \item Residual Connections: Neural network with many (deep) layers are known to learn slowly due to effects such as vanishing gradients. By bypassing (skipping) layers and connecting directly to the next (or more ahead), residual connections allow gradients to flow through the network more effectively. This can be done by skipping one or more layers of the network. 
\end{itemize}

% TODO elaborate on the framework for network connectivity with dropout and residual...

While research has studied dropout and residual connections extensively, they are often treated as two separate parameters. We propose in this research to study their synergistic impact on models and to formulate a framework where these two parameters together influence the formation of connectivity within a neural network. 

\subsection{Experimental Design}

In experiments, we will explore different combinations of dropout parameters in the transformer architecture involving various components: 1) Attention and MLP (multi-layer perceptron) layers, the number of layers skipped by residual connections, and dropout on the residual connections. Each of these parameters can have an impact on model training, convergence, and validation errors. Here are a few aspects we will investigate: 

\begin{enumerate}
    \item Dropout on Attention/MLP Layers: Lower dropout rates of 0.1 or less can be beneficial for models not prone to overfitting by allowing more weights to pass through the network for faster convergence. Higher dropout rate, e.g. greater than 0.3, can be useful for models that tend to overfit. However, the higher rate can undermine the ability to learn and converge. 
    \item Residual Skip Layers: We will investigate the standard technique to skip one layer in the residual connections to mitigate vanishing gradients. When we start to add residual connections that skip 2 or more layers, it will potentially capture longer-range dependencies, reduce complexity, but, at same time, disrupt the normal flow of gradients and make the process less stable. 
    \item Dropout on Residual Connections: While it's common to apply no dropout on residuals, we will introduce some randomness in the skip connections by introducing a dropout rate here. In this experiment, we utilized the built-in PyTorch function nn.Dropout() to apply dropout. We modified the typical implementation by eliminating the scaling factor $1 / (1 - p)$. This can further regularize the model but disturb the flow of gradients. There can be a tradeoff between regularization and stability of the model. 
\end{enumerate}

With the above parameter configurations, we will be able to train models and evaluate how the training process respond to the combinations of parameters. In general, we expect medium/lower dropout on attention/MLP layers combined with even lower dropouts on residual connections will be optimal for model training. Yet we have yet to understand how they interact and how training will be impacted when we increase the dropout on residual connections, proportionate to the number of layers they skip. There can be a potential balancing effect between learning and regularization with increasing dropout in deeper layers. 

\subsection{Benchmark Datasets}
%  Description of the Shakespeare dataset
%  Description of an additional dataset
%  Description of Evaluation metrics

We conduct our experiments on the Tiny Shakespeare, which includes 40,000 lines of text from Shakespeare's plays. This dataset has a total of 202,651 words and a vocabulary size of 25,670. 

\subsection{Software and Hardware Setup}
For this research, we implement our models in python version 3.11 with cuda v 12.3. We conduct experiments on a NVIDIA GeForce RTX 2080 ti GPU with 11 gigabyte memory and 4352 cuda cores.

\subsection{Evaluation Metrics}
% Jerry TODO: Briefly describe the evaluation metrics (# iterations, validation error, training time, etc.) in one paragraph or two. Describe what goes through in each iteration, the loss function for the error, and how is it validated. 
In the experiments, the dataset was split into two parts: 90\% for training and 10\% for validation. During the training process, the model processed batches of training data, making predictions at each step. The accuracy of these predictions was assessed using a cross-entropy loss function, a standard measure in classification tasks that evaluates the difference between the model's predicted probabilities and the actual outcomes\cite{zhang2018generalized}. 

% $$Loss = -\sum_{c=1}^My_{o,c}\log(p_{o,c})$$ % Comment out for now to save space as it's well known

Validation of the model was conducted using the 10\% reserved dataset, where the model's performance on unseen data was evaluated using the same loss function. This step is to ensure the model is learning general patterns and not just memorizing the training data (overfitting).

\begin{itemize}
    \item Number of iterations: Number of iterations refers to the total count of times the learning algorithm processes the training data, with each iteration (a mini batch) updating the model's parameters based on the loss function.
    \item Validation error: Validation error measures the model's accuracy, based on cross-entropy loss, on a separate, unseen subset of data, serving as an indicator of its ability to generalize beyond the training set.
    \item Training time: Training time is the duration it takes for a machine learning model to complete its training on the given dataset, influenced by factors like dataset size, model complexity, and computational resources. 
\end{itemize}

\section{Experiments}

Initial experiments show a significant interaction between the dropout rates in attention and residual connections, affecting both the speed of training convergence and the validation error metrics. % The outcomes of this research will offer practical insights into optimizing transformer models for language tasks, potentially improving training processes and accuracy in natural language processing applications.

\subsection{Baseline Dropout Results}
% TODO: Present the initial results before optimization.
% TODO: Discuss the shortcomings in these initial experiments. 

For baseline result, we have no dropout in attention and residual connections. Model includes 16 layers, 4 attention heads, and an embedding size of 128. Training utilizes a batch size of 8 and a block size of 64 characters, providing a context of up to 64 previous characters for each prediction. The learning rate is initially set at 1e-3, with provisions for decay to a minimum of 1e-4 across 50,000 iterations.

\begin{figure}[h]
  \centering
  \begin{minipage}{0.5\linewidth}
    \centering
    \includegraphics[width=1.0\linewidth]{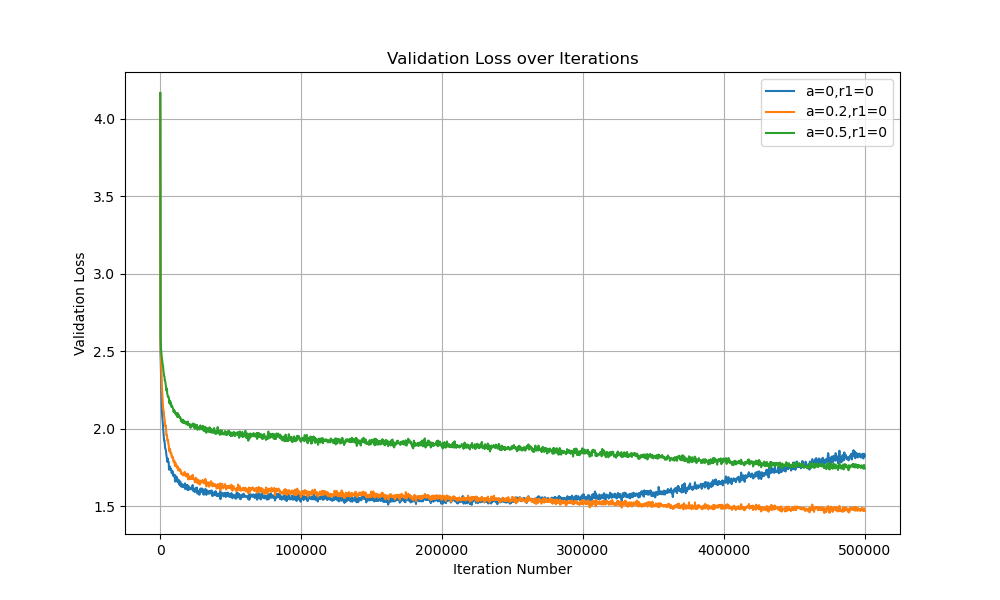}\\
    (a)
  \end{minipage}%
  \begin{minipage}{0.5\linewidth}
    \centering
    \includegraphics[width=1.0\linewidth]{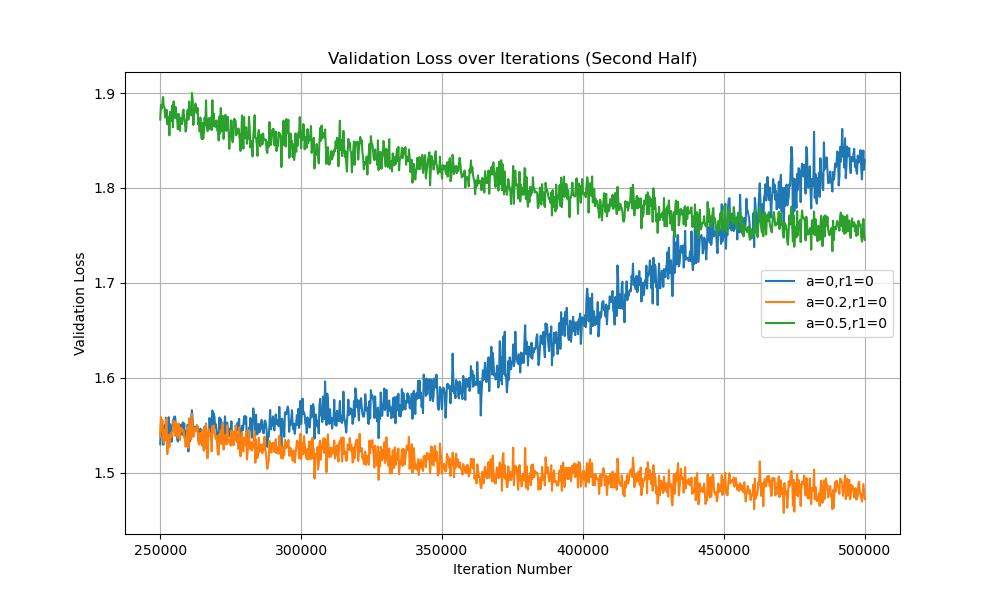}
    (b)
  \end{minipage}
  \caption{Attention Dropout}
  \label{fig:baseline}
\end{figure}

% Jerry TODO: 
% Please include the figure and results on attention/MLP dropout without residual as the baseline. 

In our experiment on attention connection, as shown in Figure~\ref{fig:baseline}, we conducted a series of test with a fixed dropout rate of 0 for the residual connections, while varying the dropout rate for attention and MLP connections. We found that a dropout rate below 0.1 for the attention mechanism leads to overfitting issues within the model (blue line with 0 dropout). Conversely, implementing a relatively high dropout rate for attention necessitates a longer convergence period for the model (green line with 0.5 dropout).

% Jerry TODO: 
% 1. Overlay a cutout of iterations after 15000 on top to show; 
% 2. Legend: remove "validation error" (no need to repeat), and put "a=0, r1=0" etc. 
% 3. Y axis should be labeled "validation error"
% 4. Change title to "Validation Loss over Training Iterations"

% Discuss the finding and consistency with previous research regarding an optimal dropout rate here. 

% Jerry TODO: 
% For each of the following, please include dropout results for each skip-n residual connection, while holding previous parameter constant/optimal. 
\subsection{Dropout on Skip-1 Residual}

\begin{figure}[h]
  \centering
  \begin{minipage}{0.5\linewidth}
    \centering
    \includegraphics[width=1.0\linewidth]{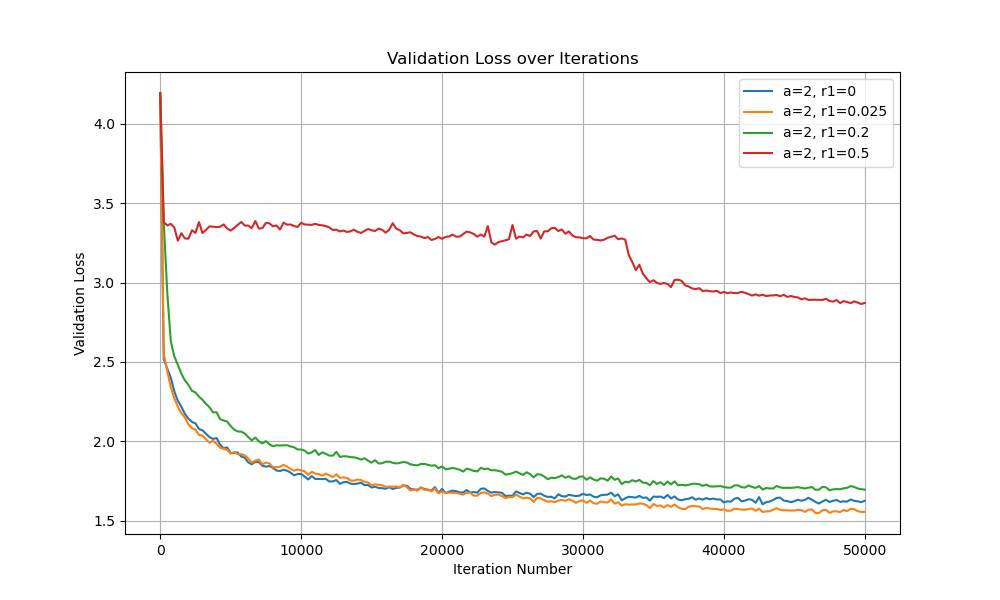}
    (a)
  \end{minipage}%
  \begin{minipage}{0.5\linewidth}
    \centering
    \includegraphics[width=1.0\linewidth]{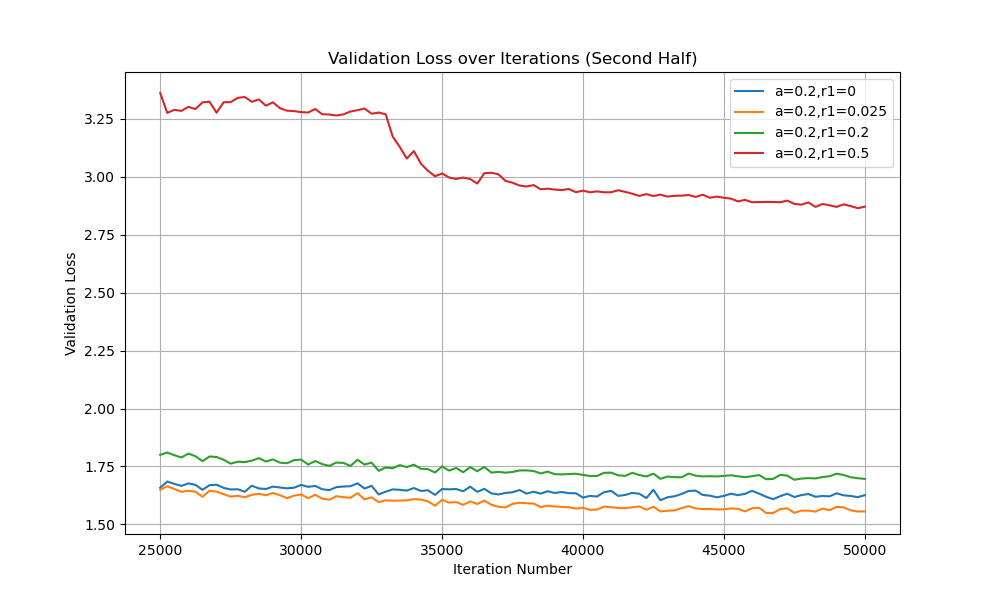}
    (b)
  \end{minipage}
  \caption{Skip-1 Residual Dropout}
  \label{fig:Skip-1 Residual}
\end{figure}

In our experiment on single-layer skip connection, we adjusted the dropout rates for the residual connections while maintaining dropout rate for the attention connections 0.2. As shown in Figure ~\ref{fig:Skip-1 Residual}, we observed that a lower dropout rate for skip-1-layer residual connections improved model performance. Meanwhile, a higher dropout rate applied to 1 layer residual connections complicates the model's ability to converge.

\subsection{Dropout on Skip-2 Residual}

\begin{figure}[h]
  \centering
  \begin{minipage}{0.5\linewidth}
    \centering
    \includegraphics[width=1.0\linewidth]{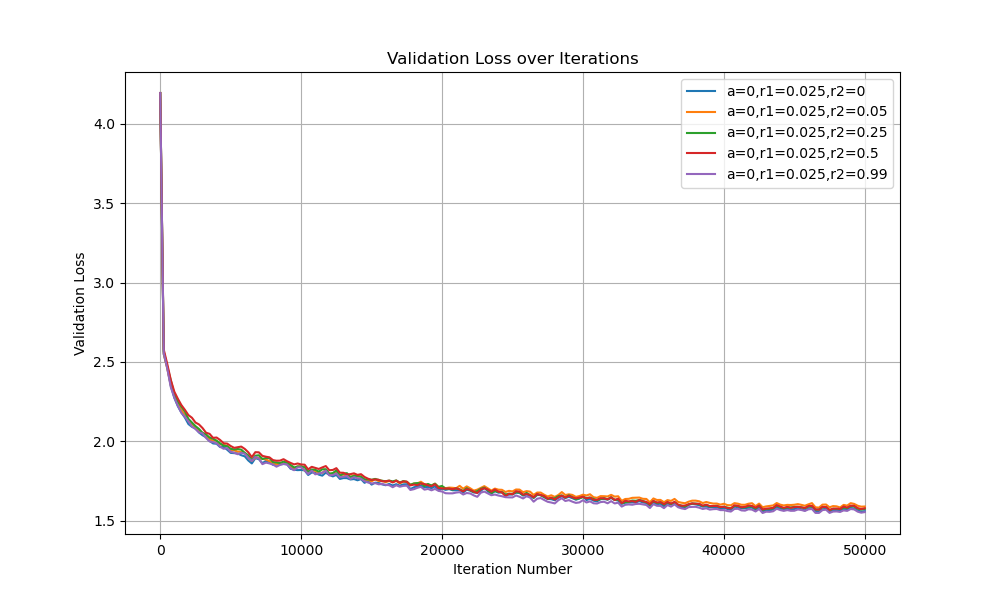}
    (a)
  \end{minipage}%
  \begin{minipage}{0.5\linewidth}
    \centering
    \includegraphics[width=1.0\linewidth]{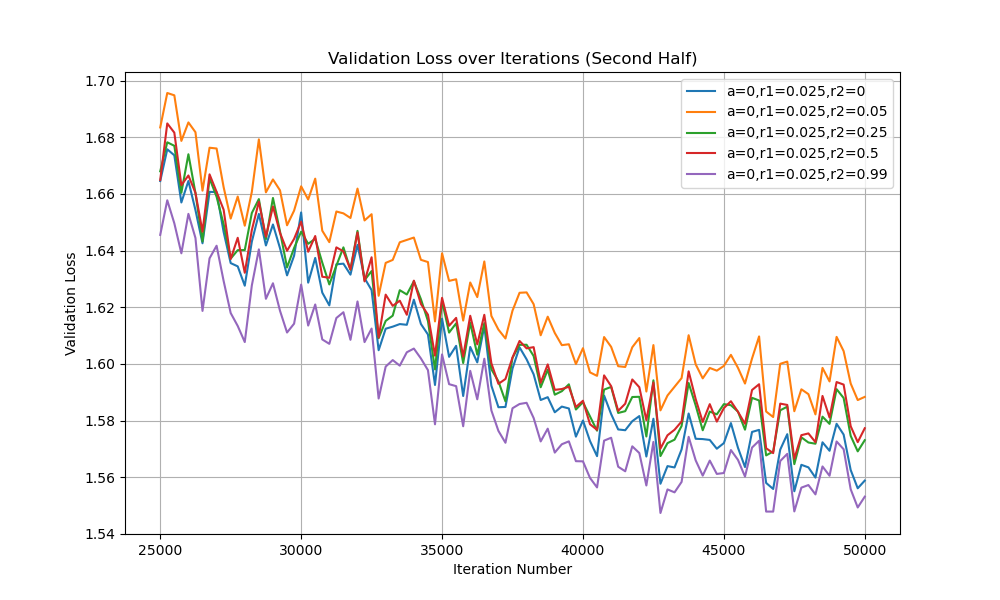}
    (b)
  \end{minipage}
  \caption{Skip-2 Residual Dropout}
  \label{fig:Skip-2 Residual}
\end{figure}

In our experiment on a network with two-layer skip connections, we manipulated the dropout rates applied to the two-layer residual connections while keeping the dropout rate for the attention connections constant at 0.2 and for the single-layer residual dropout at 0.025. As shown in Figure ~\ref{fig:Skip-2 Residual}, we found that the model exhibited enhanced performance when a relatively high dropout rate, such as 0.99, was applied to the two-layer residual connections. From these results, we infer that not all residual connections contribute positively to the model's performance. Given the 16 layers of the entire model, it appears that skip-2-layer residual connections have limited benefits and should be applied conservatively. 

\subsection{Dropout on Skip-4 Residual}

In our experiment on a network with four-layer skip connections, we manipulated the dropout rates applied to the four-layer residual connections while keeping the dropout rate for the attention connections constant at 0.2 and for the single-layer residual connection at 0.025, and two-layer residual connection at 0.99. In this situation, we found that adding more residual connection did not help the training process.

\begin{figure}[h]
  \centering
  \begin{minipage}{0.5\linewidth}
    \centering
    \includegraphics[width=1.0\linewidth]{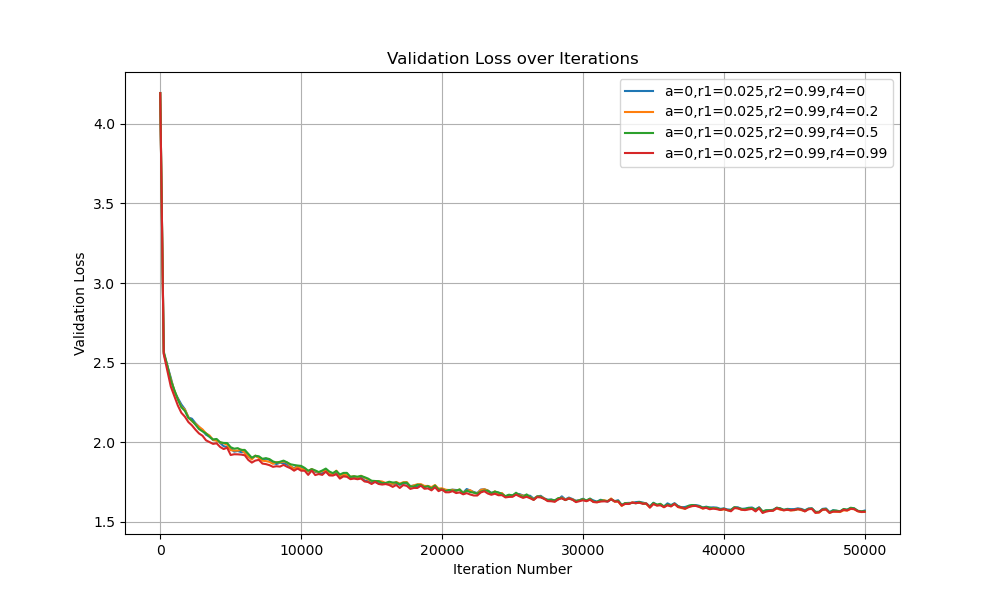}
    (a)
  \end{minipage}%
  \begin{minipage}{0.5\linewidth}
    \centering
    \includegraphics[width=1.0\linewidth]{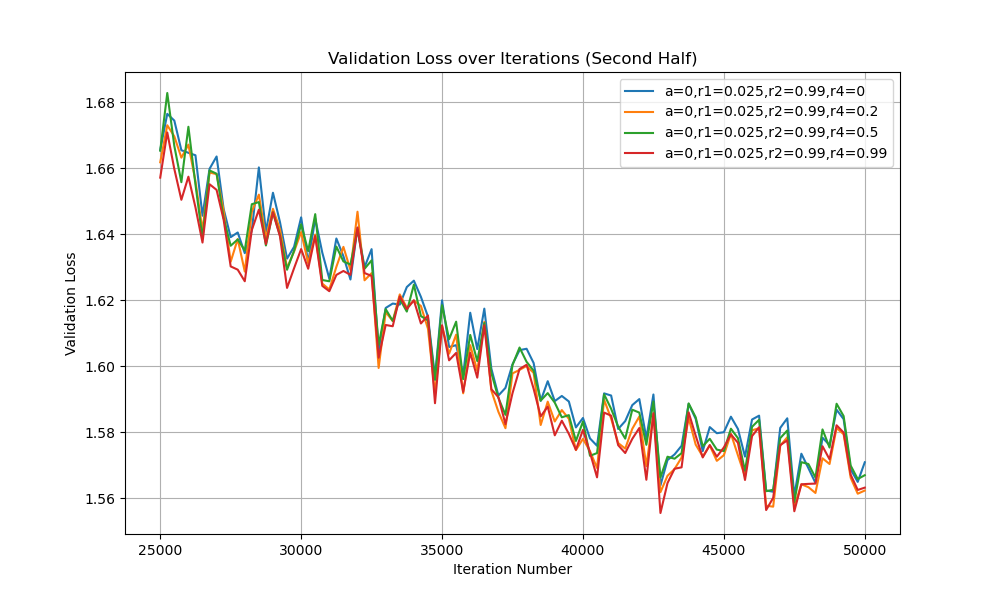}
    (b)
  \end{minipage}
  \caption{Skip-4 Residual Dropout}
  \label{fig:Skip-4 Residual}
\end{figure}

We then varied the dropout rates for these four-layer residual connections, again maintaining the same parameter values for previous layers. The results are shown in Figure ~\ref{fig:Skip-4 Residual}. Our observations indicated that the addition of more residual connections, specifically the four-layer ones, in fact degraded the performances. This outcome suggests that the effectiveness of incorporating additional layers of residual connections may have diminishing returns in terms of influencing the network's training dynamics and overall performance. Determining the depth of the skip connections, combined with a dropout, will likely depend on the depth and complexity of the entire neural network. 

\subsection{Final Comparison}

In our experiments, we delved into the effects of dropout across various configurations of residual connection layers. Our baseline involved a 0.2 dropout rate on attention and MLP, with no dropout on residual connections, yielding a loss of 1.6261 after 50,000 iterations as shown in Table~\ref{tab:overall}. Upon introducing a 0.025 dropout rate to a single-layer residual connection, we observed an improved loss of 1.5560, indicating a better outcome compared to the no-dropout single residual connection network.

Further experimentation with a two-layer residual connection, utilizing an optimal dropout rate of 0.99, yielded a loss of 1.5531, subtly surpassing the results from the single-layer residual connection. However, expanding the network to include four layers of residual connections and optimizing the dropout rate for this configuration led to a loss of 1.5624. This result was slightly worse to two-layer residual connection, showing that adding skip-4-layer residuals is not beneficial for the model of 16 layers total. 

\begin{table}
\centering
\label{table:data}
\begin{tabular}{
  c % Attn
  c % Res1
  c % Res2
  c % Res4
  c % Iters
  c % Loss
}
\toprule
\textbf{Attn} & \textbf{Res1} & \textbf{Res2} & \textbf{Res4} & \textbf{Iters} & \textbf{Loss} \\
\midrule
0    & 0     & {N/A} & {N/A} & 50000 & 1.5618 \\
\textbf{0.2} & \textbf{0}    & \textbf{N/A} & \textbf{N/A} & \textbf{50000} & \textbf{1.6261} \\
0.5  & 0     & {N/A} & {N/A} & 50000 & 1.9533 \\
0.2  & 0     & {N/A} & {N/A} & 50000 & 1.6261 \\
\textbf{0.2}  & \textbf{0.025} & \textbf{N/A} & \textbf{N/A} & \textbf{50000} & \textbf{1.5560} \\
0.2  & 0.2   & {N/A} & {N/A} & 50000 & 1.6961 \\
0.2  & 0.5   & {N/A} & {N/A} & 50000 & 2.8716 \\
0.2  & 0.025 & 0     & {N/A} & 50000 & 1.5588 \\
0.2  & 0.025 & 0.05  & {N/A} & 50000 & 1.5883 \\
0.2  & 0.025 & 0.25  & {N/A} & 50000 & 1.5731 \\
0.2  & 0.025 & 0.5   & {N/A} & 50000 & 1.5773 \\
\textbf{0.2}  & \textbf{0.025} & \textbf{0.99}  & \textbf{N/A} & \textbf{50000} & \textbf{1.5531} \\
0.2  & 0.025 & 0.99  & 0     & 50000 & 1.5709 \\
0.2  & 0.025 & 0.99  & 0.2   & 50000 & 1.5624 \\
0.2  & 0.025 & 0.99  & 0.5   & 50000 & 1.5670 \\
0.2  & 0.025 & 0.99  & 0.99  & 50000 & 1.5633 \\
\bottomrule
\end{tabular}
\caption{Overall Results}
\label{tab:overall}
\end{table}

% \section{Discussion}
% TODO: Analyze the results in the context of question answering and information retrieval.
% TODO: Discuss any limitations and potential for future work.

\section{Conclusion}
% TODO: Summarize the key findings and their implications.
% TODO: Suggest future directions for research in this area.

% TODO The following is a hypothetical analysis, yet to be confirmed by results. 

There is indeed a trade-off between regularization (dropout), smooth learning (residual) and stability (residual side effect) of the language model. Each combination of these parameters leads to different training dynamics. Lower dropout rates generally speed up convergence but increase the risk of overfitting, while higher rates regularize the model more strongly at the cost of slower learning. While residual connections mitigate vanishing gradients and improve information flow in the network, there is a risk in disrupting the normal flow and making the model unstable. Our results show the interaction between these factors and shed light on optimal combinations of related parameters for model training and generalization. 

The results in this study are preliminary and the differences among the various configurations are marginal. However, we have only explored a relatively small model on a small dataset and plan to scale up the experiments with significantly deeper neural networks where these optimization techniques will become more critical. The minor trade-off we observed in this study can potentially become much more significant.

% The number of layers skipped by residual connections and the application of dropout on these connections are more experimental and could lead to unpredictable effects on training stability and model performance. Therefore, careful experimentation and validation are crucial to identify the optimal configuration for a given task and dataset. 
% 

%%
%% The acknowledgments section is defined using the "acks" environment
%% (and NOT an unnumbered section). This ensures the proper
%% identification of the section in the article metadata, and the
%% consistent spelling of the heading.
% \begin{acks}
% % TODO: Acknowledge any assistance or contributions from others.

% \end{acks}

\clearpage
%%
%% The next two lines define the bibliography style to be used, and
%% the bibliography file.
% TODO: List all the references in `references.bib` and use \cite{key} inline. 
\bibliographystyle{ACM-Reference-Format}
\bibliography{main}

%%% -*-BibTeX-*-
%%% Do NOT edit. File created by BibTeX with style
%%% ACM-Reference-Format-Journals [18-Jan-2012].

\begin{thebibliography}{7}

%%% ====================================================================
%%% NOTE TO THE USER: you can override these defaults by providing
%%% customized versions of any of these macros before the \bibliography
%%% command.  Each of them MUST provide its own final punctuation,
%%% except for \shownote{}, \showDOI{}, and \showURL{}.  The latter two
%%% do not use final punctuation, in order to avoid confusing it with
%%% the Web address.
%%%
%%% To suppress output of a particular field, define its macro to expand
%%% to an empty string, or better, \unskip, like this:
%%%
%%% \newcommand{\showDOI}[1]{\unskip}   % LaTeX syntax
%%%
%%% \def \showDOI #1{\unskip}           % plain TeX syntax
%%%
%%% ====================================================================

\ifx \showCODEN    \undefined \def \showCODEN     #1{\unskip}     \fi
\ifx \showDOI      \undefined \def \showDOI       #1{#1}\fi
\ifx \showISBNx    \undefined \def \showISBNx     #1{\unskip}     \fi
\ifx \showISBNxiii \undefined \def \showISBNxiii  #1{\unskip}     \fi
\ifx \showISSN     \undefined \def \showISSN      #1{\unskip}     \fi
\ifx \showLCCN     \undefined \def \showLCCN      #1{\unskip}     \fi
\ifx \shownote     \undefined \def \shownote      #1{#1}          \fi
\ifx \showarticletitle \undefined \def \showarticletitle #1{#1}   \fi
\ifx \showURL      \undefined \def \showURL       {\relax}        \fi
% The following commands are used for tagged output and should be
% invisible to TeX
\providecommand\bibfield[2]{#2}
\providecommand\bibinfo[2]{#2}
\providecommand\natexlab[1]{#1}
\providecommand\showeprint[2][]{arXiv:#2}

\bibitem[Ba et~al\mbox{.}(2016)]%
        {ba2016layer}
\bibfield{author}{\bibinfo{person}{Jimmy~Lei Ba}, \bibinfo{person}{Jamie~Ryan
  Kiros}, {and} \bibinfo{person}{Geoffrey~E Hinton}.}
  \bibinfo{year}{2016}\natexlab{}.
\newblock \showarticletitle{Layer normalization}.
\newblock \bibinfo{journal}{\emph{arXiv preprint arXiv:1607.06450}}
  (\bibinfo{year}{2016}).
\newblock


\bibitem[He et~al\mbox{.}(2016)]%
        {he2016deep}
\bibfield{author}{\bibinfo{person}{Kaiming He}, \bibinfo{person}{Xiangyu
  Zhang}, \bibinfo{person}{Shaoqing Ren}, {and} \bibinfo{person}{Jian Sun}.}
  \bibinfo{year}{2016}\natexlab{}.
\newblock \showarticletitle{Deep residual learning for image recognition}. In
  \bibinfo{booktitle}{\emph{Proceedings of the IEEE conference on computer
  vision and pattern recognition}}. \bibinfo{pages}{770--778}.
\newblock


\bibitem[LeCun et~al\mbox{.}(2015)]%
        {lecun2015deep}
\bibfield{author}{\bibinfo{person}{Yann LeCun}, \bibinfo{person}{Yoshua
  Bengio}, {and} \bibinfo{person}{Geoffrey Hinton}.}
  \bibinfo{year}{2015}\natexlab{}.
\newblock \showarticletitle{Deep learning}.
\newblock \bibinfo{journal}{\emph{nature}} \bibinfo{volume}{521},
  \bibinfo{number}{7553} (\bibinfo{year}{2015}), \bibinfo{pages}{436--444}.
\newblock


\bibitem[Rosenfeld(2000)]%
        {rosenfeld2000two}
\bibfield{author}{\bibinfo{person}{Ronald Rosenfeld}.}
  \bibinfo{year}{2000}\natexlab{}.
\newblock \showarticletitle{Two decades of statistical language modeling: Where
  do we go from here?}
\newblock \bibinfo{journal}{\emph{Proc. IEEE}} \bibinfo{volume}{88},
  \bibinfo{number}{8} (\bibinfo{year}{2000}), \bibinfo{pages}{1270--1278}.
\newblock


\bibitem[Srivastava et~al\mbox{.}(2014)]%
        {srivastava2014dropout}
\bibfield{author}{\bibinfo{person}{Nitish Srivastava},
  \bibinfo{person}{Geoffrey Hinton}, \bibinfo{person}{Alex Krizhevsky},
  \bibinfo{person}{Ilya Sutskever}, {and} \bibinfo{person}{Ruslan
  Salakhutdinov}.} \bibinfo{year}{2014}\natexlab{}.
\newblock \showarticletitle{Dropout: a simple way to prevent neural networks
  from overfitting}.
\newblock \bibinfo{journal}{\emph{The journal of machine learning research}}
  \bibinfo{volume}{15}, \bibinfo{number}{1} (\bibinfo{year}{2014}),
  \bibinfo{pages}{1929--1958}.
\newblock


\bibitem[Vaswani et~al\mbox{.}(2017)]%
        {vaswani2017attention}
\bibfield{author}{\bibinfo{person}{Ashish Vaswani}, \bibinfo{person}{Noam
  Shazeer}, \bibinfo{person}{Niki Parmar}, \bibinfo{person}{Jakob Uszkoreit},
  \bibinfo{person}{Llion Jones}, \bibinfo{person}{Aidan~N Gomez},
  \bibinfo{person}{{\L}ukasz Kaiser}, {and} \bibinfo{person}{Illia
  Polosukhin}.} \bibinfo{year}{2017}\natexlab{}.
\newblock \showarticletitle{Attention is all you need}.
\newblock \bibinfo{journal}{\emph{Advances in neural information processing
  systems}}  \bibinfo{volume}{30} (\bibinfo{year}{2017}).
\newblock


\bibitem[Zhang and Sabuncu(2018)]%
        {zhang2018generalized}
\bibfield{author}{\bibinfo{person}{Zhilu Zhang} {and} \bibinfo{person}{Mert
  Sabuncu}.} \bibinfo{year}{2018}\natexlab{}.
\newblock \showarticletitle{Generalized cross entropy loss for training deep
  neural networks with noisy labels}.
\newblock \bibinfo{journal}{\emph{Advances in neural information processing
  systems}}  \bibinfo{volume}{31} (\bibinfo{year}{2018}).
\newblock


\end{thebibliography}

%%
%% If your work has an appendix, this is the place to put it.
% \appendix

\end{document}